\documentclass[a4paper,conference]{IEEEtran}
\ifCLASSINFOpdf
\else
\fi
\usepackage{amsmath}

\usepackage{graphicx,color}

\usepackage{adjustbox}

\usepackage[english]{babel} 

\usepackage{paralist} 

\usepackage{booktabs} 

\usepackage{color} 

\usepackage{gensymb}

\usepackage{algorithm}

\usepackage{algorithmic}

\usepackage{tabularx,booktabs}

\newcommand\newnotecommand[3]{%
\newcommand#1[1]{{\color{#3}\footnote{{\color{#3}#2:} ##1}}}}
\newnotecommand\joni{Joni}{red}
\newnotecommand\wenyan{Wenyan}{yellow}
\newnotecommand\cory{Cory}{blue}
\newnotecommand\fatemeh{Fatemeh}{green}

\begin{document}

\title{Object Detection in Equirectangular Panorama}

\author{\IEEEauthorblockN{Wenyan Yang, Yanlin Qian, Joni-Kristian K{\"a}m{\"a}r{\"a}inen}
\IEEEauthorblockA{Laboratory of Signal Processing\\
Tampere University of Technology, Finland\\
\texttt{First.Surname@tut.fi}}
\and
\IEEEauthorblockN{Francesco Cricri, Lixin Fan}
\IEEEauthorblockA{Nokia Technologies\\
Tampere, Finland}}

\maketitle

\begin{abstract}
  We introduce a high-resolution equirectangular panorama (aka 360-degree,
  virtual reality, VR) dataset for object detection and propose a
  multi-projection variant of the YOLO detector.
  The main challenges with
  equirectangular panorama images are i) the lack of annotated training
  data, ii) high-resolution imagery 
  and iii) severe geometric distortions of objects near
  the panorama projection poles.
  In this work, we solve the challenges by I) using training examples available 
  in the ``conventional datasets'' (ImageNet and COCO), II) employing only
  low resolution images that require only moderate
  GPU computing power and memory, and III) our multi-projection YOLO handles
  projection distortions by making multiple stereographic
  sub-projections.
  In our experiments, YOLO outperforms the other state-of-the-art
  detector, Faster R-CNN, and our
  multi-projection YOLO achieves the best accuracy with low-resolution
  input.
\end{abstract}


%
\IEEEpeerreviewmaketitle

\section{Introduction}
\label{sec:intro}
\begin{figure}[t] 
  \begin{center}
    \includegraphics[width=0.49\linewidth,height=0.22\linewidth]{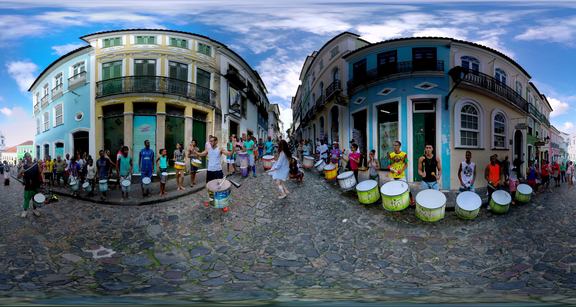}
    \includegraphics[width=0.49\linewidth,height=0.22\linewidth]{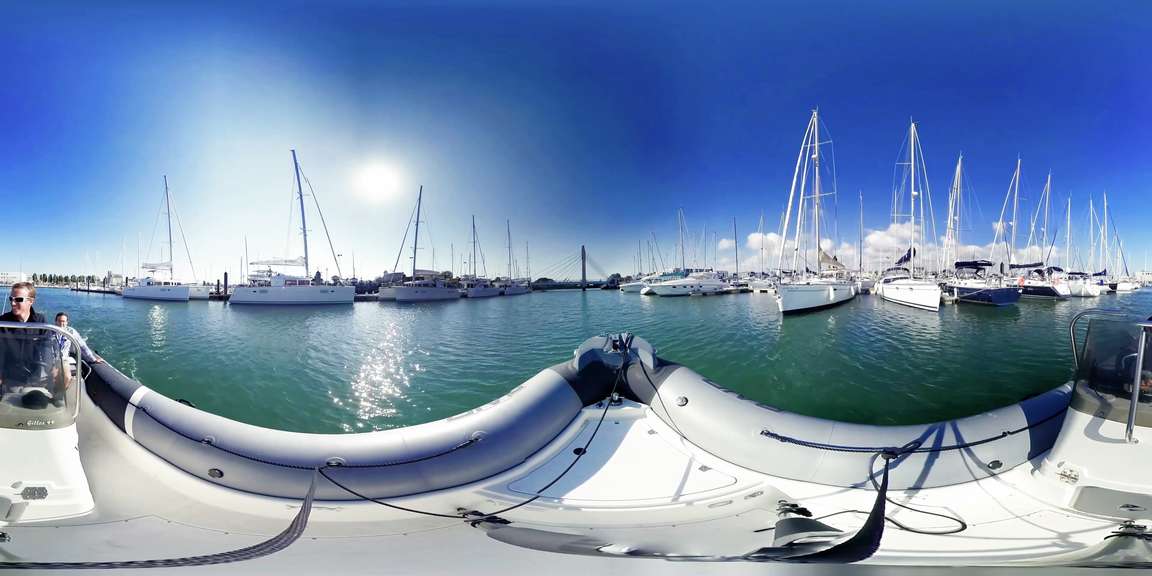}\\

    \vspace{0.5\medskipamount}
    
    \includegraphics[width=0.49\linewidth,height=0.22\linewidth]{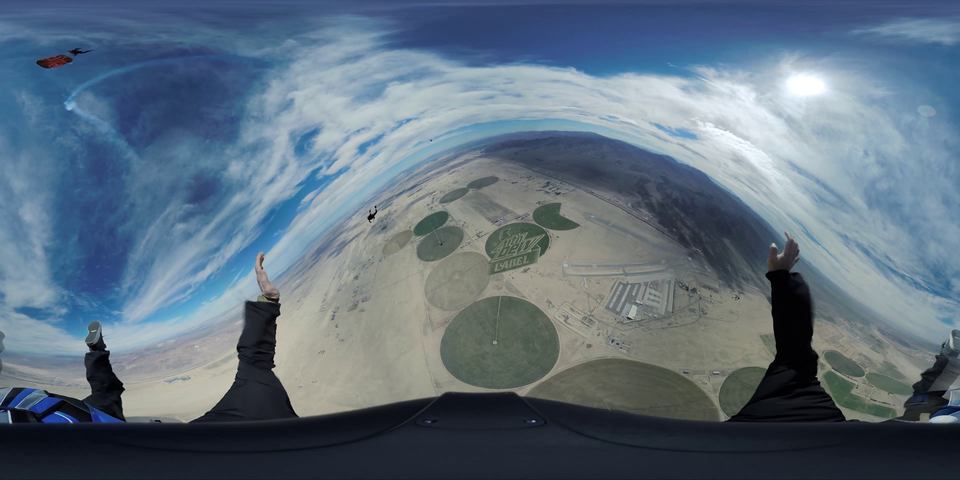}
    \includegraphics[width=0.49\linewidth,height=0.22\linewidth]{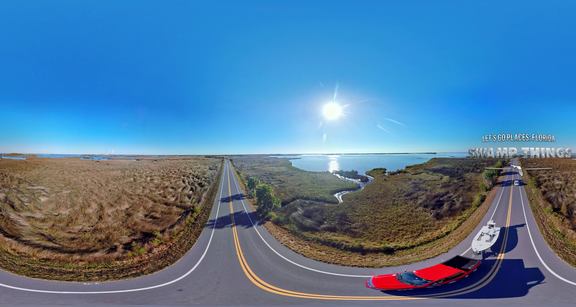}\\

    \caption{Frames from 360$^{\circ}$ videos captured by users and uploaded to Youtube (distortions are due to
      the equirectangular panorama projection). In the bottom-left the projection horizon does not match
      to the scene horizon and needs to be fixed (automatically done in our dataset).}
    \label{fig:examples}
  \end{center}
\end{figure}

360-degree ($360^{\circ}$) video and image content has recently gained momentum
due to wide availability of consumer-level video capture and display devices
- ``Virtual Reality (VR) gear''. Equirectangular panorama
(ERA, Figure~\ref{fig:examples}) has quickly become the main format to store
and transmit VR video. ERA images create new challenges
for computer vision and image processing as i) we lack annotated 360 datasets
for many problems, ii) imagery
are often of high-resolution to cover the viewing sphere with reasonable resolution and
iii) equirectangular projection creates severe geometric distortions for objects
away from the central horizontal line. 

In computer vision community, there are several recent works on processing
360 video and images, for example, ``compression'' of wide angle VR video
to conventional narrow angle video~\cite{Su-2016-accv,Su-2017-cvpr},
equirectangular super-resolution~\cite{FakourSevom-2018-visapp} and
360-degree object tracking~\cite{Kart-2018-visapp}. In this work, we focus on visual
object class detection in equirectangular panorama and provide
a novel 360-degree dataset and evaluate state-of-the-art object
detectors in this novel problem setting.

We train two state-of-the-art detectors, Faster R-CNN~\cite{Ren-2015-nips} and
YOLO (version 2)~\cite{Redmon-2017-cvpr}, with conventional examples available
in the existing datasets (ImageNet and COCO in our case) and test them with
360-degree data. In our experiments, the YOLO detector performs better
than Faster R-CNN, but for dense scenes of many
objects YOLO needs high-resolution input (Sec.~\ref{sec:resolution}). To adapt the YOLO detector for less computation power,
we propose a multi-projection variant of the original YOLO detector. Our
m-p YOLO employs stereographic projection and post-processing with
soft non-maximum suppression (soft-NMS) and outperforms both Faster R-CNN
and YOLO. Our novel contributions in this work are: 
\begin{compactitem}
\item We introduce a high-resolution equirectangular panorama dataset for
  evaluating and benchmarking object detectors with 360-degree data (Sec.~\ref{sec:dataset}).
\item We compare two state-of-the-art object detectors,  
  Faster R-CNN~\cite{Ren-2015-nips} and YOLO (version 2)~\cite{Redmon-2017-cvpr}
  trained with conventional data (ImageNet and COCO) in 360-degree object detection
  (Sec.~\ref{sec:experiments}).
\item We propose a multi-projection variant of YOLO that achieves high accuracy with
  moderate GPU power (Sec.~\ref{sec:method}).
\end{compactitem}
Our dataset and code will be made publicly available.




\section{Equirectangular Panorama Dataset}
\label{sec:dataset}
360$^{\circ}$ virtual reality images and video frames are stored
as 2D projections of the captured 3D world on the surface of
a {\em viewing sphere}.  The most popular projection is
{\em equirectangular panorama} (ERA) in which each sphere point is
uniquely defined by two angles~\cite{Snyder-1987-book}:
{\em latitude} $\varphi \in[-90^\circ,+90^\circ]$ and
{\em longitude} $\lambda \in[-180^\circ,+180^\circ]$. In this work,
we employ equirectangular panorama images.

For our dataset we selected 22 4k-resolution ($3840 \times 1920$)
VR videos captured and uploaded by users to Youtube. From
each video we selected a number of frames and annotated a set of
visual classes in each frame. Our dataset consists of the total of 903
frames and 7,199 annotated objects. For the experiments, we
selected objects that are also available 
in the Microsoft COCO dataset~\cite{Lin-2014-eccv} to experiment
training with examples in conventional images. The selected videos
represent dynamic activities, for example, skateboarding or riding
bicycles, are all real world scenes, and contain both moving and static objects.

\subsection{Object Annotations}
\label{sec:annotation}

\begin{figure}[h] 
  \begin{center}
    \includegraphics[width=0.98\linewidth]{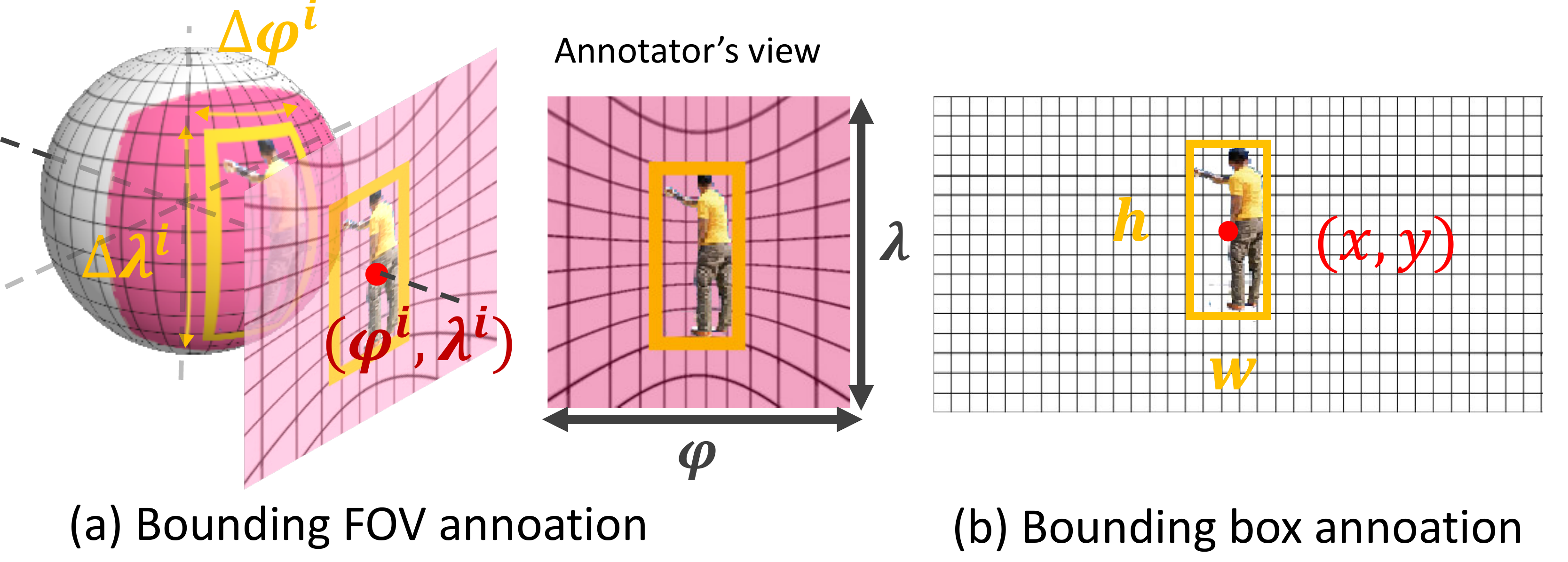}\\
    \caption{Example of our annotation protocol: a) user centers the object and 
      annotates ``Bounding FOV'' (BFOV) in center and FOV angles. b) BBox (bounding box) coordinates
      are directly annotated on the original panorama.
      \label{fig:two_types}}
  \end{center}
\end{figure}

An important difference between equirectangular panorama and
conventional images is that the
panorama projection changes object's appearance depending on its
spatial location (Figure~\ref{fig:examples}). Therefore for consistent annotation,
we need to select a ``canonical pose'' where bounding box coordinates are valid and
has the shape of a box. We implemented an annotation tool where the user is shown
a perspective projection of the full panorama. User is allowed to change the center
of projection and field-of-view (FOV). During the annotation workshop,
annotators were asked to move an object to the central view where the object center
approximately matches the projection axis and then annotate a bounding box.
As for the ground truth we store the following attributes: object label $l^i$; the bounding box center as
angular coordinates $\varphi^i$, $\lambda^i$; and the bounding box angular
dimensions $\Delta\varphi^i$, $\Delta\lambda^i$. We refer such annotation as BFOV (Boundig FOV, see Figure~\ref{fig:two_types}).

\begin{figure}[h] 
  \begin{center}
    \includegraphics[width=1.0\linewidth,height=0.45\linewidth]{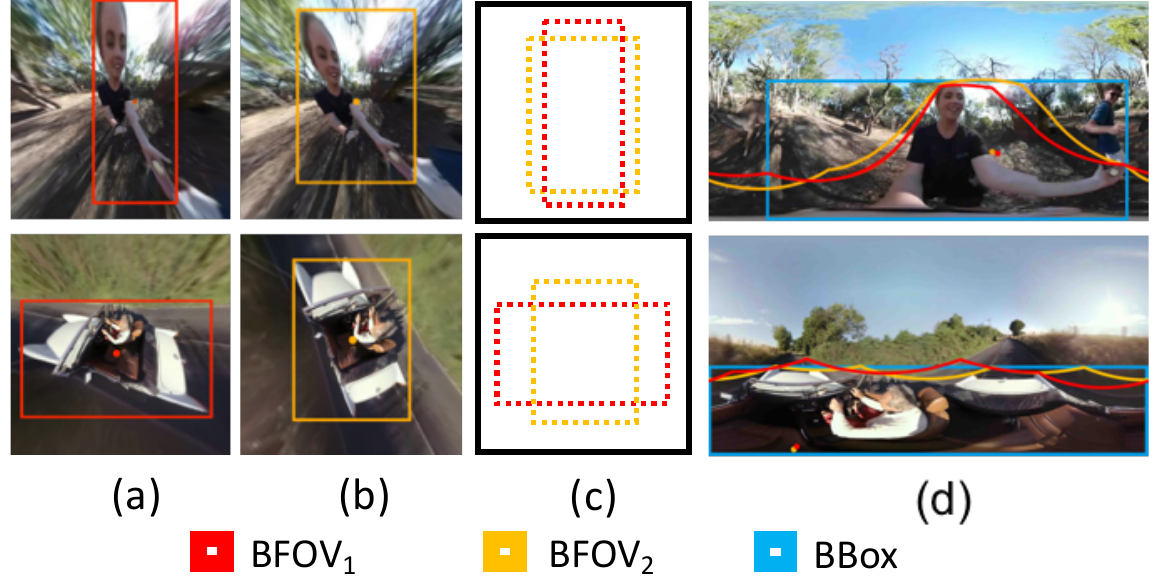}\\
    \caption{Example of annotation ambiguities of objects near the VR camera.
      (a) and (b) show bounding boxes annotated by two different annotators who both set annotation
      tool field-of-view to $\varphi=150^\circ$, $\lambda=150^\circ$. (c) illustrates
      bounding box ambiguity due to sensitivity to the bounding box center that defines the projection
      angle (cf. (a) and (b)). (d) illustrates the annotated bbs in the ERA image and blue box
      is the corrected bounding box.
      \label{fig:distort_examples}}
  \end{center}
\end{figure}

It should be noted that the above annotation protocol has problems with
objects that are close to the VR camera. Annotating these objects requires a
very wide FOV ($\ge 120^\circ$) which makes annotation sensitive to the selection
of the projection center. See Figure~\ref{fig:distort_examples} for examples of
the annotation ambiguity problem with objects too close to the camera. We manually
identified these objects and re-annotated them directly in the ERA image.
Hu et al.~\cite{hu2017deep} avoided this problem in their recent work by selecting
images where is only one ``dominating object'', but in our dataset images contain multiple
objects which are all annotated. All experiments are conducted using the original
or corrected bounding boxes in the original equirectangular panorama frames.

\subsection{360-degree dataset vs. COCO dataset}
\begin{figure}[h] 
  \begin{center}
    \includegraphics[width=0.65\linewidth]{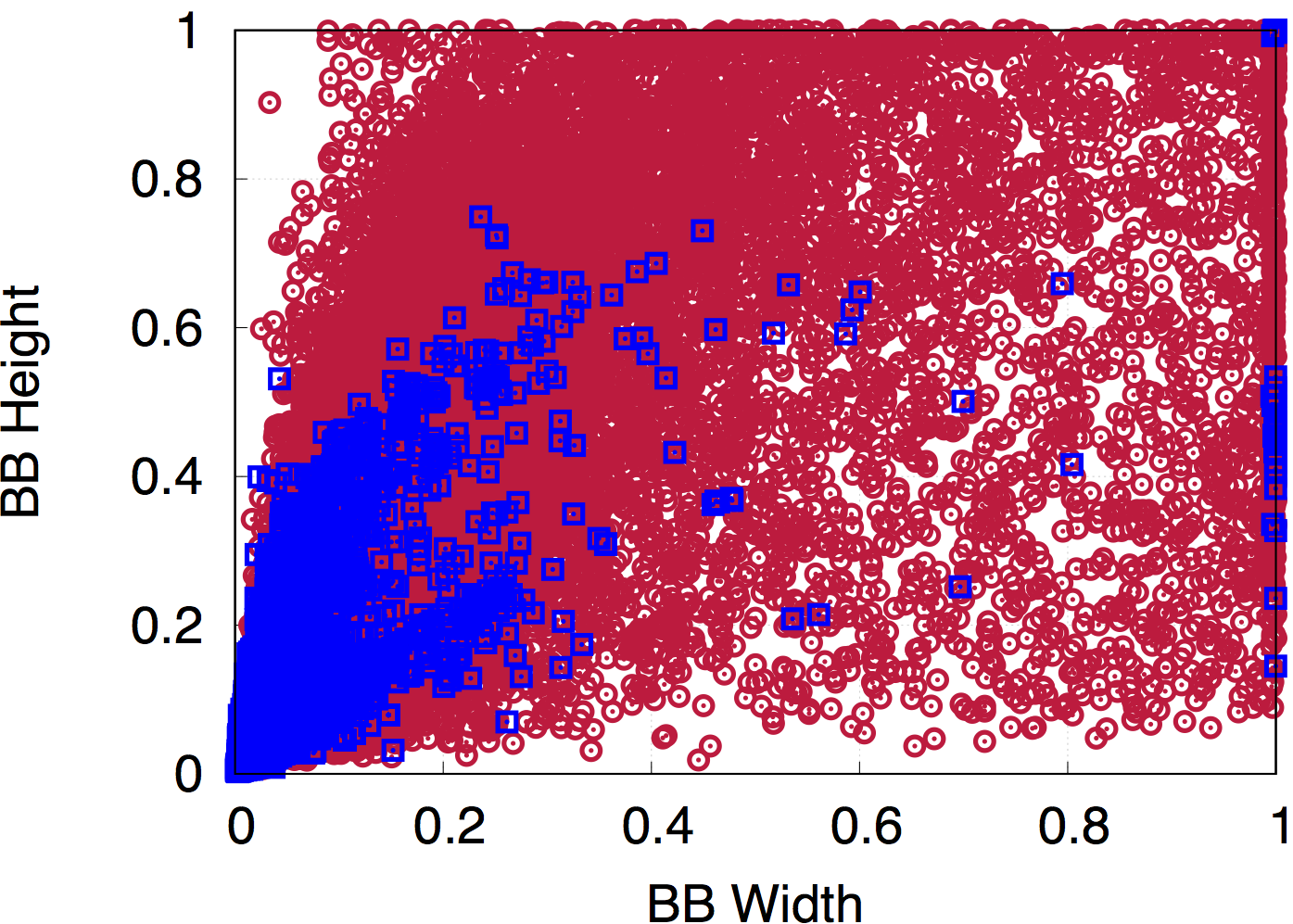} 
    \caption{Bounding box aspect ratios (normalized width and height) in the COCO (red circles) and our (blue squares) dataset.\label{fig:data_distributions}}
  \end{center}
\end{figure}

\begin{figure}[h]
  \begin{center}
    \includegraphics[width=0.8\linewidth]{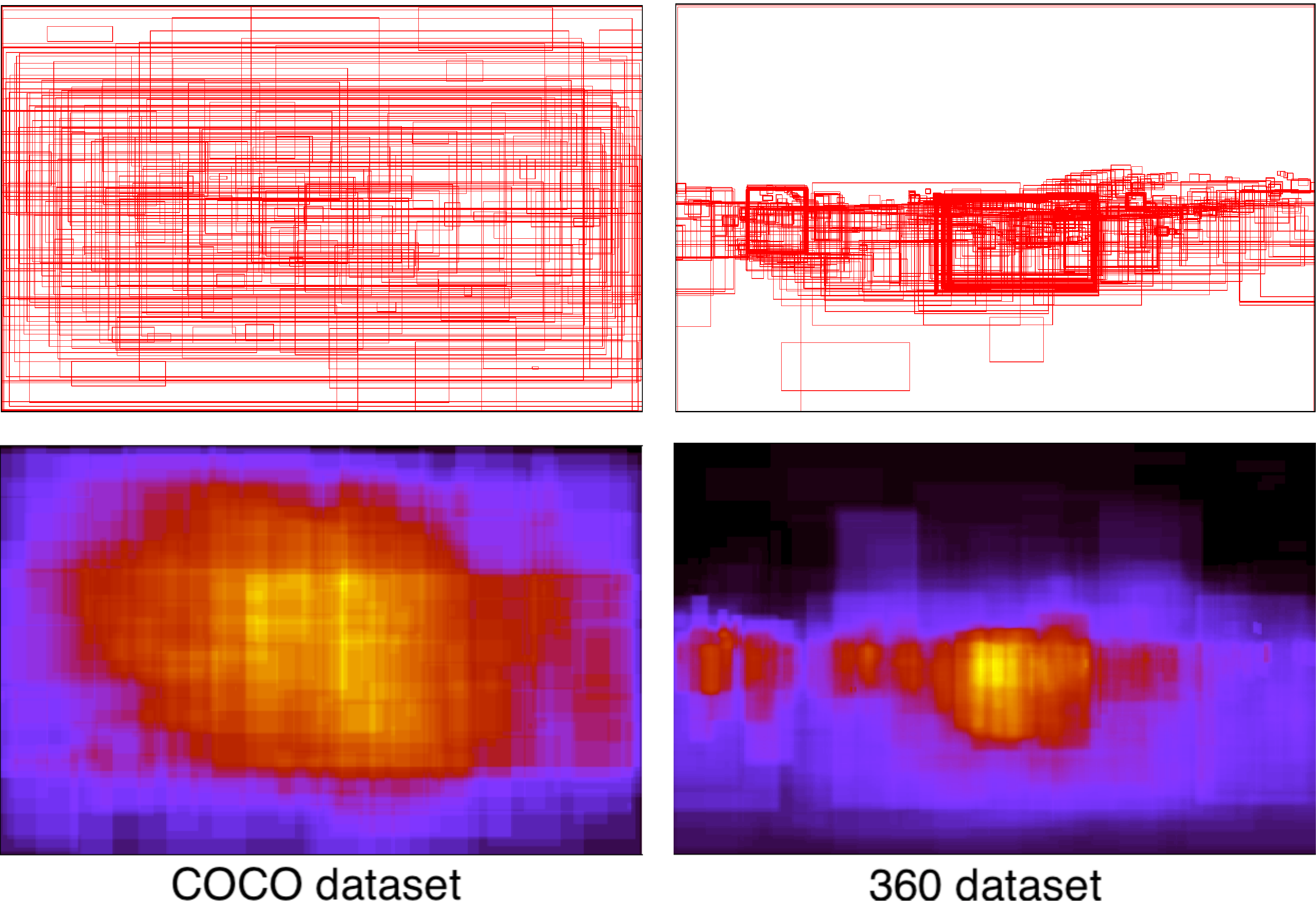}\\ 

    \caption{Bounding boxes location distribution in COCO (left) and our dataset (right) plotted to a single image. Heat maps below.\label{fig:bb}}
  \end{center}
\end{figure}

One target in our experiments is to leverage the pre-trained detectors (YOLO~\cite{Redmon-2017-cvpr} and Faster R-CNN ~\cite{Ren-2015-nips} on Microsoft COCO~\cite{Lin-2014-eccv} and ILSVRC2012~\cite{Russakovsky-2015-ijcv}) on equirectangular images. 
However, it is unclear how well object appearance match between the VR and conventional datasets.
Our annotated bounding boxes in our and COCO datasets are plotted in Figure~\ref{fig:data_distributions} and
Figure~\ref{fig:bb}. BBox dimensions (width and height normalized to
$[0,1]$) are plotted in Figure~\ref{fig:data_distributions} to compare the
aspect ratios of annotated bounding boxes. Figure~\ref{fig:bb} shows how BBoxes are distributed on images.
It is clear that COCO bounding box aspect ratios span wider range than
our dataset and locations of bounding boxes for our datasets are more limited.
This verifies that appearance of COCO objects match to sufficient degree the appearance of objects
in our dataset (except objects near the VR camera).


The main differences in our dataset as compared to COCO can be summarized as:
\begin{compactitem}
\item In our dataset there are more small objects. This can be explained by the nature of
  equirectangular panorama projection where scale quickly changes with respect to object distance. Therefore
  accurate object detection also requires high-resolution. This is evident in
  Figure~\ref{fig:data_distributions} where our objects span clearly smaller areas in images.
\item Geometric distortion due to equirectangular projection. However, in many cases people prefer
  to capture main targets near the central horizontal line where distortions are small
  if the objects are not too close to the VR camera (Figure~\ref{fig:bb}).
\end{compactitem}

For our experiments we selected $6,431$ annotated objects that are also available in the COCO
dataset: {\em person} (3720), {\em car} (1610), {\em boat} (963). Another annotated classes are
skateboard, bicycle, truck, sign, ball, penguin and lion, but since a) there were not enough
annotated objects in them or b) they are not available in COCO these were omitted in our
experiments.

\section{Multi-projection YOLO}
\label{sec:method}

In our experiments we report results for the recent state-of-the-art
object detector YOLO~\cite{Redmon-2017-cvpr} which is pre-trained on ImageNet data and
fine-tuned with COCO. However, processing full panorama with sufficient accuracy requires
high-resolution input. It requires high-end GPU (e.g. TITAN-X) to fit all data to memory and therefore 
we propose a multi-projection YOLO detector. Multi-projection YOLO can be computed on
consumer GPUs and provides competitive accuracy.

As discussed in Section~\ref{sec:annotation}, 
One of the main challenges of object detection in
equirectangular panorama is severe geometric distortions caused by panorama projection. 
A straightforward solution to remove distortions is to project sub-widows of
the full 360-degree sphere onto 2D image plane for which detection methods trained on
conventional images should perform well. However, the smaller is the sub-window field-of-view
(FOV) the more windows are needed and therefore more processing is needed.
Some recent works argue that this approach is not optimal~\cite{khasanova2017graph,su2017flat2sphere},
but methods using re-projections have also been proposed~\cite{Carroll-2009-siggraph}. 

\begin{figure*}
\centering
\includegraphics[width=0.9\textwidth]{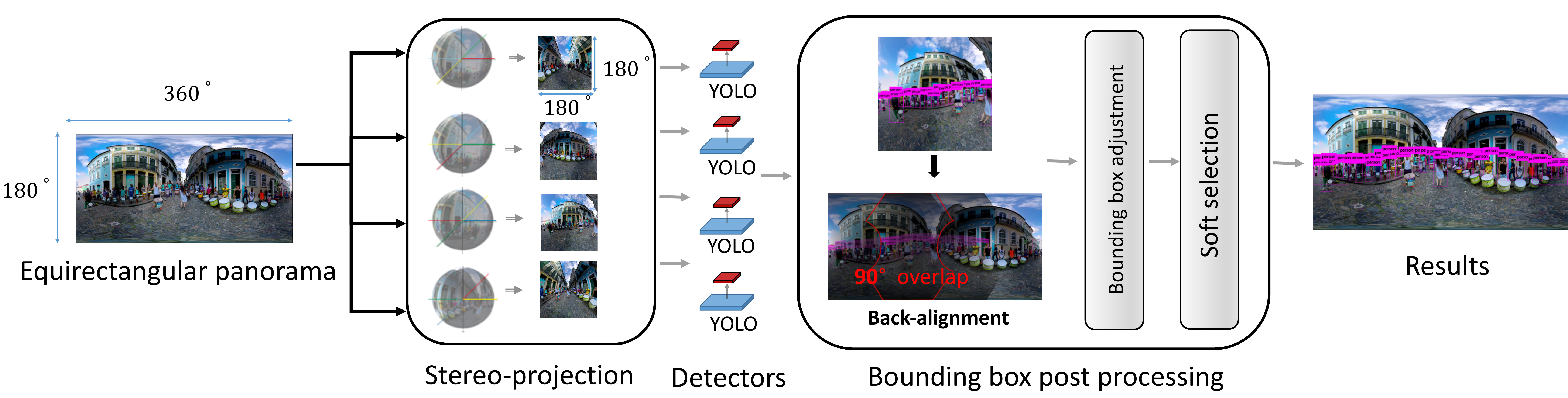}
\caption{Overall processing pipeline of our multi-projection YOLO. In the first step, we
  generate four stereographic projections for which their horizontal and vertical spans are $180^\circ$.
  The horizontal overlap between two neighbor projections is $90^\circ$. Then each sub-projection is
  separately processed by the YOLO detector. In the post-processing part, bounding box adjustment  part fixes distorted detection boxes (Figure~\ref{fig:alignment}) and soft-selection re-scores redundant boxes in overlapping areas (Section~\ref{sec:soft-selection})\label{fig:overview}}  
\end{figure*}

In our work, we investigate the multi-projection approach with a wide FOV and adopt
soft selection as an alternative to the standard non-maximum suppression (NMS)
to select detections produced by
multiple windows (see Figure~\ref{fig:overview} for overview of our approach). 

\subsection{Sphere-to-plane projection}
\label{sec:projections}

\begin{figure}[h]
  \begin{center}
    \includegraphics[width=0.45\linewidth]{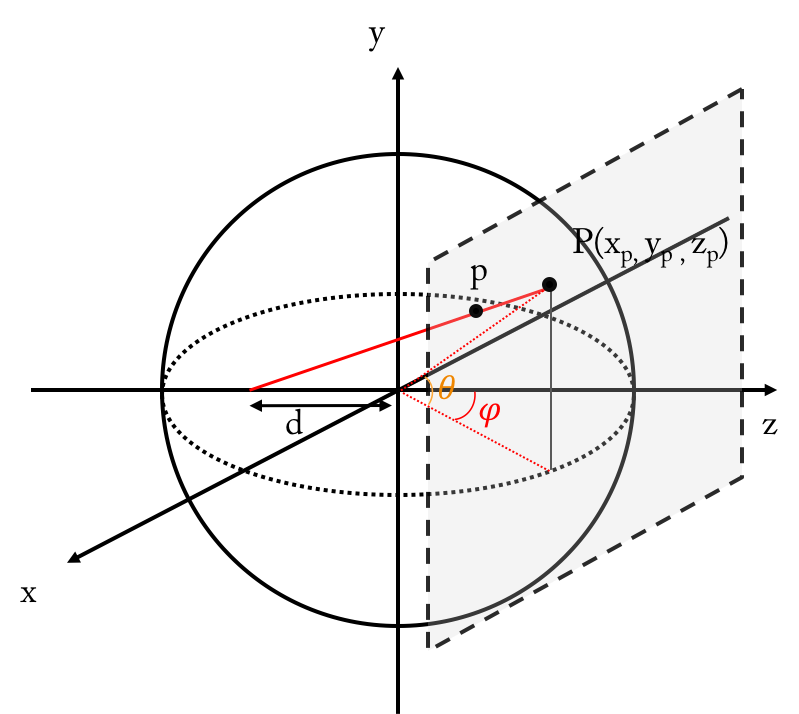}
    \includegraphics[width=0.45\linewidth]{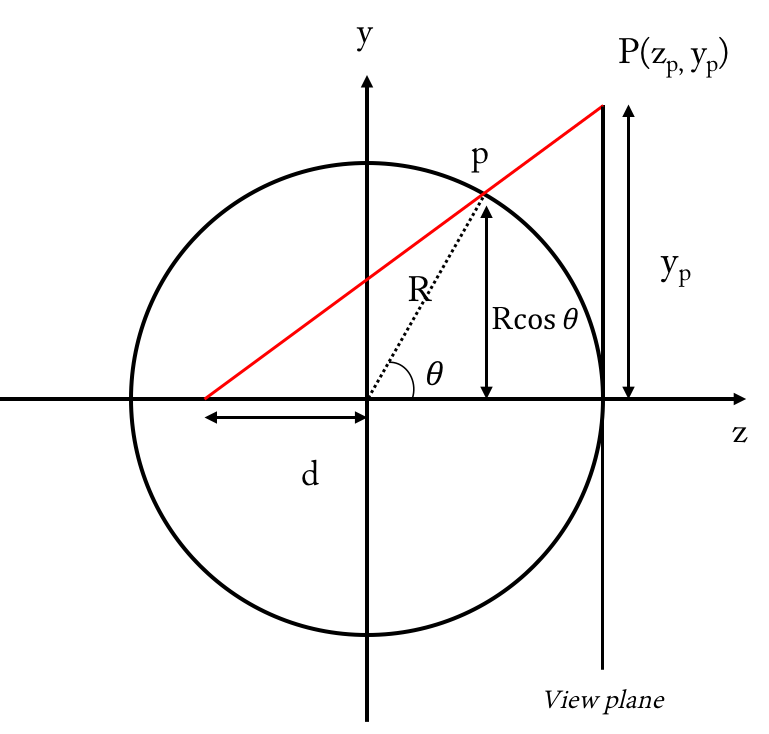}\\
    \caption{Illustration of the general sphere-to-plane projection in Eq.~(\ref{eq:projection}).
      \label{fig:projection}}
  \end{center}
\end{figure}

\begin{figure}[h]
  \begin{center}
    \includegraphics[width=0.45\linewidth,height=0.23\linewidth]{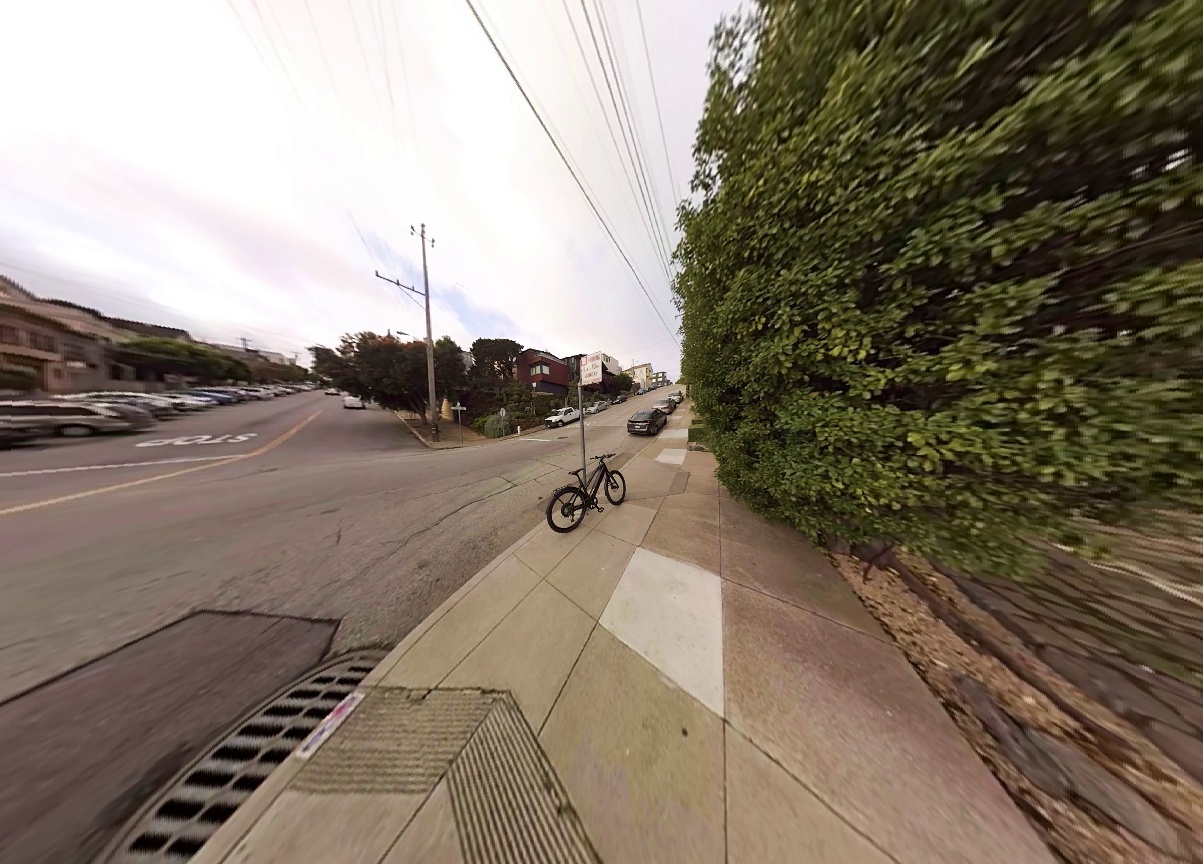}
    \includegraphics[width=0.45\linewidth,height=0.23\linewidth]{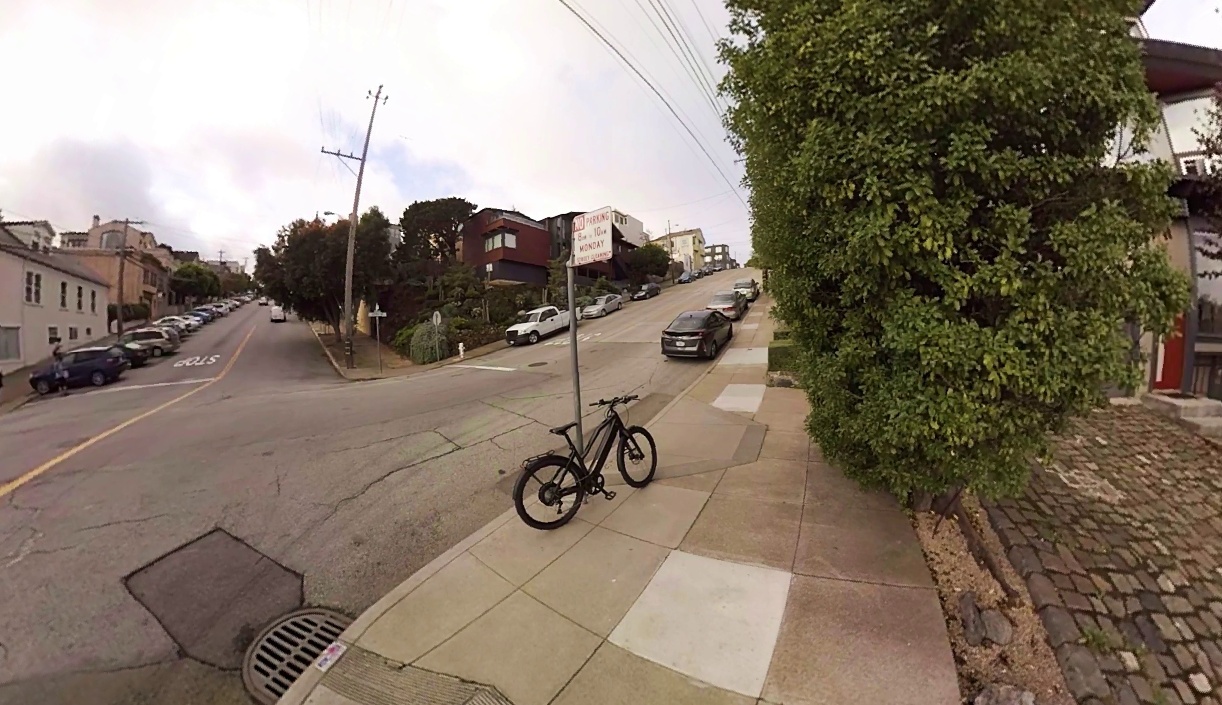}
    \caption{Perspective (left) and stereographic (right) projections of the same spherical image. The FOV for perspective projection is set as h = $90^{\circ}$, w = $175^{\circ}$ ; The FOV for stereoprojection is set as h = $90^{\circ}$, w = $180^{\circ}$. For wide-angle projection, stereographic model preserves better information although straight lines are not preserved.\label{fig:compare-projection}}
  \end{center}
\end{figure}

There are various popular projections to map a sphere to a
plane and these are mainly used in cartography applications~\cite{snyder1987map}.
The standard approach is to consider a tangent plane on a sphere and
adjust projection center and FOV in order to produce sphere-to-plane projection
(see Figure~\ref{fig:projection}). In the following
we assume that the sphere has radius $r=1$, the projection direction
is toward the positive $z$-axis, and the tangent plane center is located
at $z  =  1$. Now, a point $p(\theta,\phi)$ on the sphere is
projected onto point $P(x_p, y_p, 1)$ as
\begin{equation}
  \begin{split}
    y_p = \frac{d+1}{d + Rcos(\theta)}Rsin(\theta)\\
    x_p = \frac{d+1}{d + Rcos(\phi)}Rsin(\phi)
    \end{split}
  \label{eq:projection}
\end{equation}
If the projection center $d$ in (\ref{eq:projection}) is set to $0$,
the projection corresponds to the popular {\em perspective projection} and
for $d=1$ the projection is {\em stereographic projection}. These two
projections provide different images as illustrated in
Figure~\ref{fig:compare-projection}.

Many projection models are proposed to minimize object appearance distortions.
Chang~et al.~\cite{chang2013rectangling} proposed {\em rectangling stereo
  projection} which makes projection in two steps. At first, they project
original sphere onto a swung sphere, and then in the second step they
generate perspective projection from the swung sphere. Their method
preserves linear structures. Another distortion minimizing projection is
{\em automatic content-aware projection} by Kim et al.~\cite{kim2017automatic}.
In their work, model interpolation between local and global projections
is used to adjust distortions in optimized Pannini projection. 

In our experiments, we adopted stereographic projection model due to its simplicity. But it is easy to replace it with other models in our framework. 
\subsection{Bounding box selection}
\label{sec:proess}
\noindent\textbf{Detection -- } Two stereographic projections with horizontal and vertical
span of $180^\circ$ cover the whole sphere. However, such wide angle projections still
produce large geometric distortions for objects near the projection edges. To compensate that,
we adopt four sub-windows with overlap of $90^\circ$. In our experiments,
we use YOLOv2~\cite{Redmon-2017-cvpr} and it produces a set of object detections
with bounding boxes and detection scores for each sub-window.

\vspace{\medskipamount}\noindent\textbf{Re-aligning bounding boxes -- }
\begin{figure}[h]
  \begin{center}
    \includegraphics[width=0.9\linewidth]{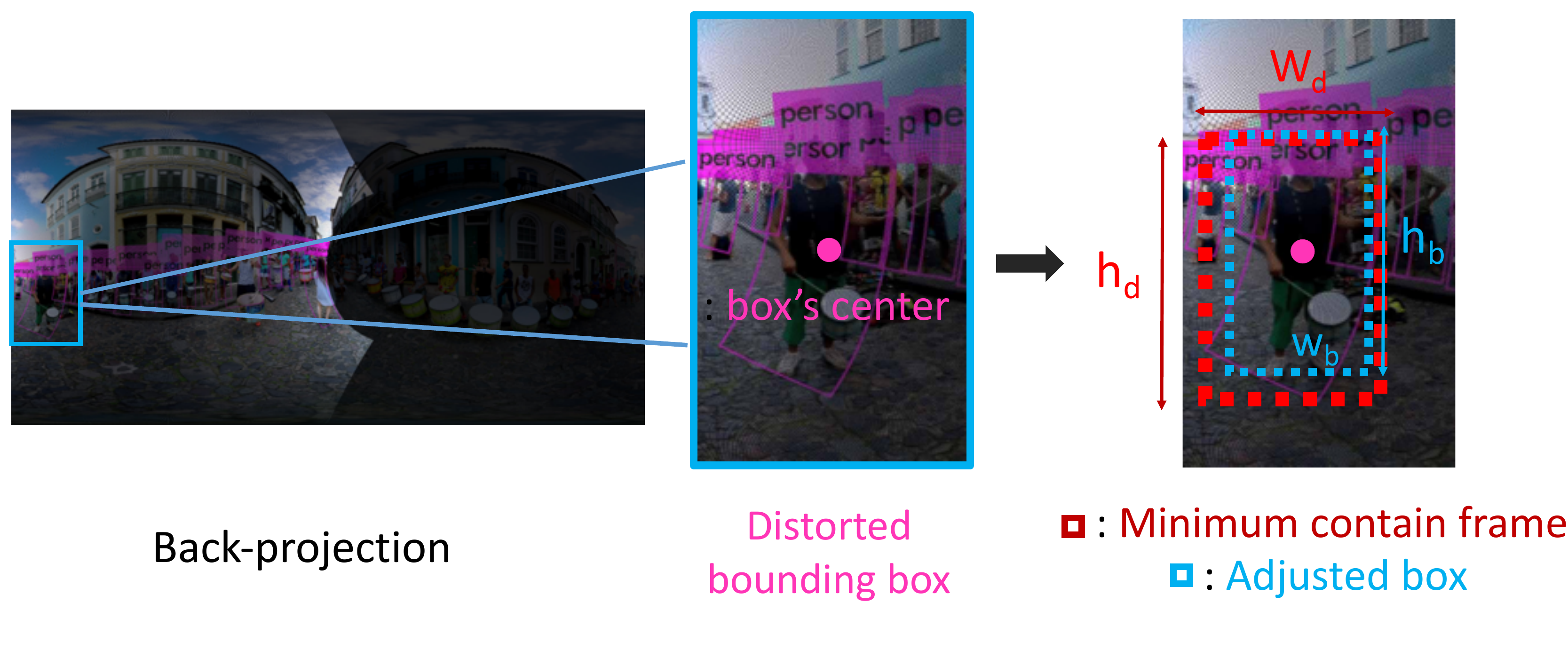}\\
    \caption{Re-aligning YOLO detector bounxing boxes. Sub-window detected bounding boxes are projected to the panorama (left) where they are distorted (middle).
      Distortions are removed by fitting a ``minimum frame'' and then using a soft penalty to fix the bounding box size. \label{fig:alignment}}
  \end{center}
\end{figure}
The YOLO detections are back-projected to the equirectangular panorama. Due to projection model's feature (Sec~\ref{sec:projections}), 
this process severely bends the lines at the edges and make evaluation difficult. Thus the BBoxes need to be re-aligned.
We exploit the fact that
bounding box center remains unaffected by the distortion and we therefore re-adjust the edges
around the center. We first find each bounding box a ``minimum frame'' that contains
the back-projected detection box. The minimum frame has its width and height $w_d$, $h_d$.
The final width and height of the adjusted box $w_b$ and $h_b$ are set based on the minimal
frame size $w_d$ and $h_d$ and the distance $d$ to the box center. Our adjustment is based
on~\cite{bodla2017soft} and the final bounding box size is set as (Fig.~\ref{fig:alignment}). The
penalty parameter is $\sigma$ and is manually set in experiments.
\begin{equation}
  w_b = {w_d}e^{-{(\frac{d^2}{\sigma}})}, \enspace
  h_b = {h_d}e^{-{(\frac{d^2}{\sigma}})} \enspace .
\end{equation}

\vspace{\medskipamount}\noindent\textbf{Soft selection -- }
\label{sec:soft-selection}
Since two neighbor sub-windows overlap by $90^\circ$ we need to post-process
the boxes in those areas. Non-maximum suppression (NMS) is the standard technique,
but yields to problems in our case since we have wide FOV detectors that are
more reliable at the center of the sub-window. Therefore, we do NMS on every
sub-window separately and for the remaining objects, we keep them all,
but re-score them based on their location in the detection window and their
overlap. This approach is inspired by~\cite{bodla2017soft}.
We do re-scoring by a soft-penalty function using the following
rule:
\begin{equation}
s'_i = {s_i}e^{-{(\frac{IoU(\hat{b},b_i)^2}{\sigma_1} + {\frac{d_i^2}{\sigma_2}}})}
\end{equation}
where $s_i$ is the original YOLO score of the bounding box detection
$b_i$ with the same label as $\hat{b}$, but futher away from the
sub-window center than $\hat{b}$. Penalty is affected by
the amount of intersection measured by Intersection-over-Union
$IoU(\hat{b},b_i)$ and the distance $d_i$ from the center of the sub-window
producing the detection. The balance between the two terms is set by
the parameters $\sigma_1$ and $\sigma_2$ and the distance $d_i$ is
normalized to $[0,1]$. Intuitively, the larger is the overlap and
further the detection is away from the detection window center the
more the score will be suppressed.

\section{Experiments}
\label{sec:experiments}
In this section, we first study performance of the two state-of-the-art detectors,
{\em YOLOv2 } by Redmon et al.~\cite{Redmon-2017-cvpr} and {\em  Faster R-CNN } by
Ren et al.~\cite{Ren-2015-nips} with various input resolutions.
In the second set of experiments we investigate our
multi-projection YOLO with low-resolution inputs.

\subsection{Settings}
In our experiments, we used the publicly available ready-to-be-used models
of YOLOv and
Faster R-CNN
as the baselines and both were optimized for the Pascal VOC data. We fine-tuned 
the pre-trained baseline models using training images of all classes in the Microsoft
COCO dataset~\cite{Lin-2014-eccv}. Faster R-CNN uses the {\em VGG-16}
network~\cite{Simonyan-2015-iclr} for classification and VGG-16 is
pre-trained using
ILSVRC2012
training and validation data.

\subsection{YOLO vs. Faster R-CNN}

\begin{table}[h]
  \caption{Average precisions of Faster R-CNN~\cite{Ren-2015-nips} and  
    YOLO (ver 2)~\cite{Redmon-2017-cvpr}. R-CNN resizes its input
    so that the smallest dim. is $600$ pixels while YOLO can process input
    of any size. Our dataset's frame size is $3840\times 1920$
    (aspect ratio $2:1$). \label{tab:rcnn_vs_yolo}}
\begin{center}
  \resizebox{1.0\linewidth}{!}{ 
  \begin{tabular}{lrrrrrr}
    \toprule
                           & Tr size           & {\em Person} & {\em Car}     & {\em Boat} &       mAP\\
    \midrule
    \multicolumn{3}{c}{\em trained with Pascal VOC} \\
    YOLOv2 $416\times 416$ & -                 &  23.63       &  24.90        &    10.60    &  19.71\\ 
    YOLOv2 $608\times 608$ & -                 &  30.94       &  26.09        &    {\bf 20.75}    &  25.93\\ 
    Faster R-CNN           & -                 &  33.13       &  25.53        &    4.33    &  21.00\\ 
    \midrule
    \multicolumn{3}{c}{\em trained with COCO} \\
    YOLOv2 $416\times 416$ & $416\times 416$   &  24.00       &  17.02       &    3.63    &  14.88\\ 
    YOLOv2 $608\times 608$ & $608\times 608$   &  {\bf 39.48}       &  {\bf 32.19}        &    19.19    &  {\bf 30.29}\\ 
    Faster R-CNN           & -                 &  27.28       &  15.90        &   13.04   &  18.74\\
    \bottomrule
  \end{tabular}}
\end{center}
\end{table}

In the first experiment, we compared the two state-of-the-art detectors,
Faster R-CNN and YOLO, to detect objects from full equirectangular panorama
input. We conducted two kind of experiments: first with the
original detectors provided by the corresponding authors and the second by
re-training the same detectors with examples in the COCO dataset. As indicated
by the results in Table~\ref{tab:rcnn_vs_yolo}
YOLOv2 always achieves better accuracy than Faster R-CNN and since
it also faster to train we selected it for the remaining experiments.

\subsection{YOLO input resolution}
\label{sec:resolution}
\begin{table}[h]
\caption{Average precisions for YOLOv2 trained and tested with inputs of various resolutions (grid size is kept constant in pixels). \label{tab:yolo_panorama}}
\begin{center}
  \resizebox{1.0\linewidth}{!}{ 
  \begin{tabular}{lrrrrrr}
    \toprule
                            & Tr size           & {\em Person} & {\em Car}  & {\em Boat} &       mAP\\
    \midrule
    YOLOv2 $~416\times 416$  & $416\times 416$   & 24.00        &      17.02 &      3.63  &      14.88\\ 
    YOLOv2 $~608\times 608$  & $416\times 416$   & 38.23        &      24.37 &      14.05 &      25.55\\
    YOLOv2 $~864\times 864$  & $416\times 416$   & 40.67        &      24.08 &      15.79 &      26.85\\
    YOLOv2 $~416\times 416$  & $608\times 608$   & 30.82        &      24.08 &      11.60 &      22.17\\ 
    YOLOv2 $~608\times 608$  & $608\times 608$   & 39.48        &      32.19 &      19.19 &      30.29\\
    YOLOv2 $~864\times 864$  & $608\times 608$   & 46.40        &      30.58 &      15.73 &      30.90\\
    YOLOv2 $~864\times 416$  & $864\times 416$   & 43.17        &      29.06 &      15.13 &      29.12\\
    YOLOv2 $1696\times 864$ & $864\times 416$   & {\bf 59.87}  & {\bf 32.52}&  {\bf19.84}& {\bf 37.41}\\
    \bottomrule
 \end{tabular}}
\end{center}
\end{table}

Since the equirectangular panorama images are of high-resolution
($3840\times1920$) we wanted to investigate the two important
parameters of the YOLO method: size of the training data (COCO)
and size of the input. During the experiments the bounding box
grid size was kept constant in pixels and therefore for larger
images YOLO is able to detect smaller examples and provide more
dense bounding box detections. The results in Table~\ref{tab:yolo_panorama}
confirm that a higher training resolution always provides better
accuracy and also higher input resolution always provides better
accuracy. The maximum training resolution with our GPU card was
$864\times 416$ after which the mini batch size is enforced to
a single image and
training becomes inefficient. By fixing the grid cell size
in pixels the YOLO improved in
detection of small objects close to each other. This is beneficial as
can be seen in Figure~\ref{fig:data_distributions} which shows how our
dataset contains a larger proportion of small objects as compared to COCO.
The boat class performs poorly as compared to other classes, but this can
be explained by the facts that many boat pictures are taken from the
annotated boat itself making it highly distorted (too close to the
main lenses and in periphery of
the capturing device) and many boats are hydrocopters which are not
well presented in COCO dataset.

\subsection{Multi-projection YOLO}

\begin{figure}[h]
  \begin{center}   
    \includegraphics[width=1.0\linewidth]{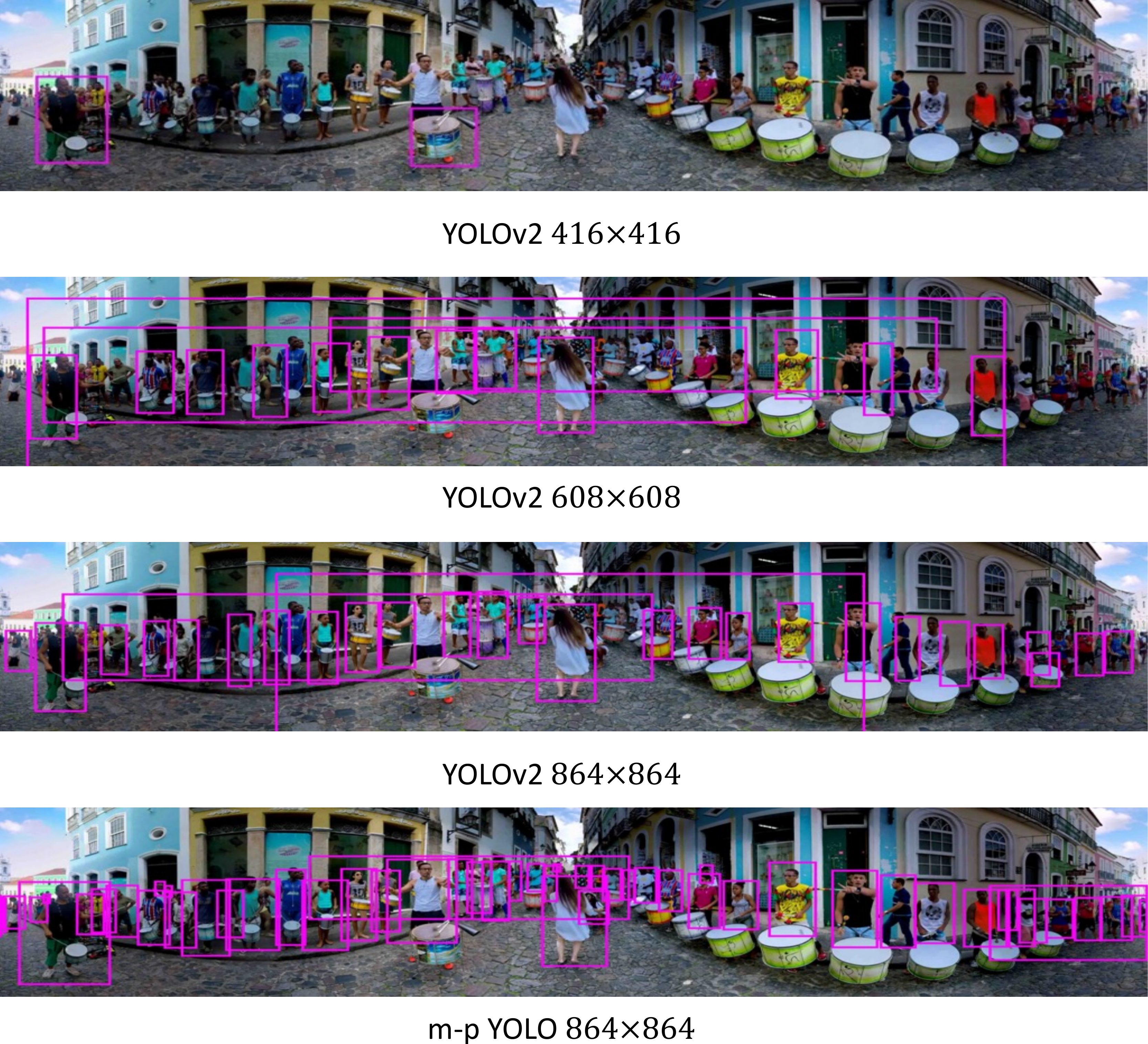}\\
    \caption{Examples of human detections with different resolutions and with multi-projection YOLO. Best viewed on display.\label{fig:compare-results}}
  \end{center}
\end{figure}

In the last experiment, we studied processing with limited GPU power in which case
we must process the input image as ``windowed'' version using our multi-projection
YOLO (Section~\ref{sec:resolution}). We fixed the input resolution to $864\times864$
and used the training resolution of $608 \times 608$. The results are shown in
Table~\ref{tab:mpYOLO} and clearly illustrate how windowed YOLO performs better than
full panorama YOLO with limited resolution (trained with the same resolution) and how
our soft-selection outperforms non-maximum suppression (NMS). Fig.~\ref{fig:compare-results}
also shows the examples of human dection. For single YOLO method, with input resolution 
incearsing the detection performs better. For windowed YOLO (m-p YOLO), it performs better
when they have the same input resolution. Although somes bounding boxes closed to edge were
enlarged due to bounding box post processing (Sec~\ref{sec:proess}).

\subsection{Ablation studies}

\begin{table}[h]
  \caption{Penalty parameters for multi-projection YOLO.\label{tab:soft}} 
  \begin{center}
    \resizebox{1.0\linewidth}{!}{ 
      \begin{tabular}{lrrrrrr}
        \toprule
        AP @ 0.5                        & {\em Person} & {\em Car}  &  {\em Boat} & {\em mAP}\\
        $\sigma_1 = 0.3, \sigma_2 = 0.3$  & 49.71 &           34.86  &      12.21 &      32.26 \\
        $\sigma_1 = 0.3, \sigma_2 = 0.6$  & {\bf54.51} &      {\bf35.42}  &      12.93 &      {\bf34.29} \\
        $\sigma_1 = 0.3, \sigma_2 = 0.9$  & 51.35 &           34.83  &      12.20 &      32.79 \\
        $\sigma_1 = 0.6, \sigma_2 = 0.3$  & 51.35 &           34.79  &      12.20 &      32.78 \\
        $\sigma_1 = 0.6, \sigma_2 = 0.6$  & 51.26 &           34.76  &      12.17 &      32.73  \\
        $\sigma_1 = 0.6, \sigma_2 = 0.9$  & 51.10 &           34.60  &      12.16 &      32.62 \\
        $\sigma_1 = 0.9, \sigma_2 = 0.3$  & 51.30 &           34.72  &      12.16 &      32.73 \\
        $\sigma_1 = 0.9, \sigma_2 = 0.6$  & 51.08 &           34.52  &      12.16 &      32.34 \\
        $\sigma_1 = 0.9, \sigma_2 = 0.9$  & 50.88 &           34.20  &      12.19 &      31.41 \\
        $NMS$                             & 47.47 &            30.96  &      12.60 &      26.31 \\
        YOLOv2 (tr. $608\times 608$)   & 46.40        &      30.58 &      {\bf15.73} &      30.90\\
        \midrule
        AP @ 0.4                        & {\em Person} & {\em Car}  &  {\em Boat} & {\em mAP}\\
        $\sigma_1 = 0.3, \sigma_2 = 0.3$  & 62.19 &           40.30  &      24.07 &      42.19 \\
        $\sigma_1 = 0.3, \sigma_2 = 0.6$  & {\bf64.07} &           40.26  &      24.05 &      42.79 \\
        $\sigma_1 = 0.3, \sigma_2 = 0.9$  & 64.03 &           42.04  &      24.05 &      {\bf43.37} \\
        $\sigma_1 = 0.6, \sigma_2 = 0.3$  & 64.03 &           40.12  &      24.01 &      42.72 \\
        $\sigma_1 = 0.6, \sigma_2 = 0.6$  & 62.24 &           {\bf43.13}  &      24.09 &      43.15  \\
        $\sigma_1 = 0.6, \sigma_2 = 0.9$  & 63.64 &           41.36  &      24.00 &      43.00 \\
        $\sigma_1 = 0.9, \sigma_2 = 0.3$  & 63.96 &           41.65  &      23.99 &      43.20 \\
        $\sigma_1 = 0.9, \sigma_2 = 0.6$  & 63.63 &           41.25  &      24.20 &      43.02 \\
        $\sigma_1 = 0.9, \sigma_2 = 0.9$  & 63.34 &           40.89  &      23.88 &      42.70 \\
        $NMS$                             & 62.66 &           38.27  &      24.43 &      41.79 \\
        YOLOv2 (tr. $608\times 608$)   & 52.27        &       38.43 &      {\bf30.41} &      40.37\\
        \bottomrule
    \end{tabular}}
  \end{center}
\end{table}

The results of the first ablation study are shown in Table~\ref{tab:soft}. In this
experiment we tested the effect of the overlapping penalty parameter $\sigma_1$ and
the distance penalty parameter $\sigma_2$. Moreover, we compared our results to
the most popular post-processing method: non-maximum suppression (NMS) with the
default threshold set to $0.3$. For all penalty term values our approach achieved better
accuracy than NMS and penalty term optimization by cross-validation resulted to
the best accuracy with $\sigma_1 = 0.3$ and $\sigma_2 = 0.6$.

\begin{table}[h]
  \caption{Our multi-projection YOLO for low-res. ($864\times 864$) input with weights trained on ($608\times 608$) size.\label{tab:mpYOLO}} 
\begin{center}
  \resizebox{1.0\linewidth}{!}{ 
  \begin{tabular}{lrrrrrr}
    \toprule
                                   & {\em Person} &  {\em Car}  &  {\em Boat} & {\em mAP}\\
    \midrule
    YOLOv2                         & 46.40        &      30.58 &     {\bf15.73} &      30.90\\
    m-p YOLO (NMS)                 &      47.47   &     30.96   &      12.60 &      26.31 \\
    m-p YOLO                       & {\bf54.51}   & {\bf35.42}  &      12.93 &      {\bf34.29} \\
    \bottomrule
  \end{tabular}}
\end{center}

\end{table}

Another important consideration with our dataset is that since bounding boxes are annotated
in central view they distort when bounding box is moved to objects' original location
(the same effect does not occur with conventional images). However, in the standard evaluation
protocol the bounding box overlap limit for successful detection is $0.5$ which is much
harder to achieve in equirectangular panorama setting. To experiment this, we tested our
penalty terms in the setting where the overlap threshold was relaxed to $0.4$. 
The results are also shown Table~\ref{tab:soft} and illustrate how the performance
improved significantly for all configurations.

\section{Conclusions}
\label{sec:conclusion}
We studied the problem of object detection in 360-degree
VR images (equirectangular panorama) and proposed a novel
benchmark dataset. Moreover, we compared two state-of-the-art
methods and showed superior performance of the recently
proposed YOLO version 2. However, good performance requires
high-resolution input (testing stage) which is not suitable
for low-end GPUs (out of memory). To compensate lower processing
power we proposed a multi-projection YOLO method that is superior
in the low-resolution setting. Our data and code will be made
publicly available.

\section*{Acknowledgement}
This work was partially funded by the Finnish Funding Agency for
Innovation (Tekes) project ``$360^\circ$ Video Intelligence - For
Research Benefit'' and 
participating companies (Nokia Technologies, Lynx Technology, JJ-Net, BigHill
Companies and Leonidas).


%



{\small
\bibliographystyle{IEEEtran}
\bibliography{object_detectors,360}
}

\end{document}